# Adversarial Attacks against Face Recognition: A Comprehensive Study

Fatemeh Vakhshiteh[a], Ahmad Nickabadi[b], Raghavendra Ramachandra[c,*]

[a]*Department of Biomedical Engineering, Amirkabir University of Technology, Tehran, Iran*
[b]*Department of Computer Engineering and Information Technology, Amirkabir University of Technology, Tehran, Iran*
[c]*Department of Information Security and Communication Technology, Norwegian Biometrics Laboratory (NBL), Norwegian University of Science and Technology (NTNU i Gjovik), Gjøvik, Norway*

**Abstract**

Face recognition (FR) systems have demonstrated outstanding verification performance, suggesting suitability for real-world applications ranging from photo tagging in social media to automated border control (ABC). In an advanced FR system with deep learning-based architecture, however, promoting the recognition efficiency alone is not sufficient, and the system should also withstand potential kinds of attacks designed to target its proficiency. Recent studies show that (deep) FR systems exhibit an intriguing vulnerability to imperceptible or perceptible but natural-looking adversarial input images that drive the model to incorrect output predictions. In this article, we present a comprehensive survey on adversarial attacks against FR systems and elaborate on the competence of new countermeasures against them. Further, we propose a taxonomy of existing attack and defense methods based on different criteria. We compare attack methods on the orientation and attributes and defense approaches on the category. Finally, we explore the challenges and potential research direction.

## 1. Introduction

Face recognition (FR) has been a prevalent biometric technique for identity authentication and is broadly used in several areas, such as finance, military, public security, and daily life. A typical FR system's ultimate goal is to identify or verify a person from a digital image or a video frame taken from a video source. Researchers describe FR as a biometric artificial intelligence-based application that can exclusively identify a person through analyzing patterns of the person's facial features.

The idea of using the face as a biometric trait inspired in the 1960s, and the design of the first successful FR system dates back to the early '90s (M. A. Turk and Pentland, 1991). In recent times, the latest advancements of deep learning, together with the use of mounting hardware and abundant data, have resulted in massive development in FR algorithms with the excellent performance (Parkhi et al., 2015; Taigman et al., 2014; Wen et al., 2016). This performance permits the broad deployment of FR technologies in further diverse applications, ranging from photo tagging in social media to dubious identification in automated border control (ABC) systems.

In an advanced FR model, however, promoting the recognition efficiency alone is not sufficient, and the system should also withstand potential kinds of attacks designed to target its proficiency. Recently, researchers found that (deep) FR systems are vulnerable against different types of attacks that create data variations to fool classifiers. These attacks can be accomplished either via (a) physical attacks, which modify the physical appearance of a face before image capturing, or (b) digital attacks, which implement modifications in the captured face image (Singh et al., 2020). Presentation attacks, also referred to as spoofing attacks (Marcel et al., 2014), are among the main techniques used for physical attacks. In contrast, adversarial attacks (Yuan et al., 2019) and the variations resulting from morphing attacks

* Corresponding author. Tel.: +47-61135458; fax: +47-61135240
  E-mail address: raghavendra.ramachandra@ntnu.no

(Scherhag et al., 2020) are critical techniques utilized for digital invasion. Note that adversarial attacks are mainly categorized in the class of digital attacks, but some methods are designed to accomplish physically.

Among different attacks, adversarial attacks are fascinating since they generally target deep neural networks (DNNs) and focus on convolutional neural networks (CNNs), based on which the state-of-the-art FR models are established. The massive growth in the number of papers published each year in the field of adversarial example generation demonstrates the attractiveness of this type of attack (see Fig. 1).

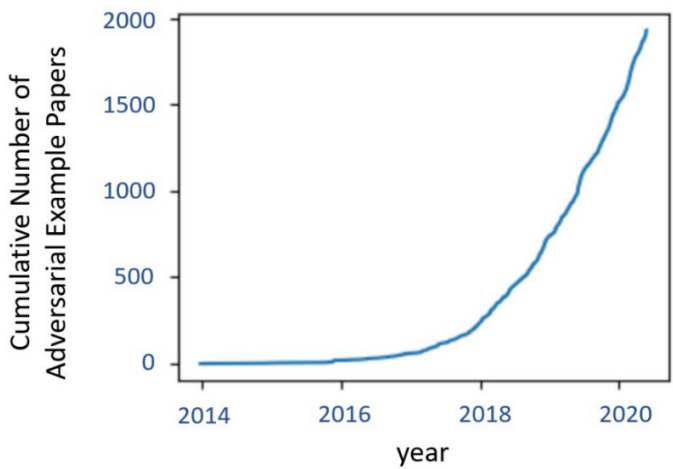

Fig. 1. The cumulative number of adversarial example papers published in recent years (Carlini, 2019).

This research presents a comprehensive survey on different techniques of adversarial attack generation intended to deceive FR systems, along with the potential countermeasures established against them. This is the first study that attempts to review adversarial attack and defense strategies on FR systems to the best of our knowledge. Since FR may refer to each of the two applications of face identification or face verification, we review both in this study.

The main contributions of this paper are:

- We review recent studies on adversarial example generation approaches on FR systems, present an illustrative taxonomy of the corresponding methods according to their orientation, and compare these approaches on orientation and attributes.
- We review the new adversarial example detection methods regarding the FR systems, categorize the presented algorithms, and demonstrate a descriptive taxonomy of this classification.
- We outline the main challenges and potential solutions for adversarial examples targeting FR models based on four main problems: Particularization/Specification of adversarial examples, instability of FR models, deviation from the human vision system, and image-agnostic perturbation generation.

The remainder of this paper is organized as follows: Section II introduces the background of FR techniques, architectures, and datasets. In Section III, we describe the standard terms related to adversarial attacks and defenses in the context of the FR course, describe the attacks' attributes, explain the experimental standards, and discuss the pioneer methods of generating attacks. We review adversarial example generation methods intended to deceive the FR mission in Section IV. We discuss the methods and compare the approaches based on orientation and attributes. In Section V, corresponding countermeasures are investigated. We discuss current challenges and potential future research directions in Section VI. Section VII concludes the work.

## 2. Background

In this section, we briefly introduce basic FR systems and elaborate on incorporated models in the era of deep learning. Next, we present widely used architectures and standard datasets in this regard.

*2.1. A Brief Introduction to Face Recognition*

Face recognition has been an age-old research topic in the computer vision community, and the first success of it dates back to the '90s. Since then, this research path has undergone scientific leaps in four decisive times, according to which face representation for recognition has taken sequential forms of holistic learning, local feature learning, shallow learning, and deep learning (Wang and Deng, 2018).

In the early 1990s, the historical Eigenface approach (M. Turk and Pentland, 1991) was introduced, and the study of FR became popular shortly after that. From then till the 2000s, the holistic approaches extracted low-dimensional representations from face images based on certain distribution assumptions (Deng et al., 2012; He et al., 2005; Moghaddam et al., 1998; Zhang et al., 2011) dominated the FR community. Nevertheless, these methods demonstrated a failure in addressing the uncontrolled facial modifications that deviate from the prior considered assumptions. In the early 2000s, local-feature-based FR techniques were introduced, and handcrafted descriptors such as Gabor (Liu and Wechsler, 2002) and LBP (Ahonen et al., 2006) became popular. However, distinctiveness and compactness were the two properties these local features lacked. In the early 2010s, local learning-based features were introduced (Cao et al., 2010; Chan et al., 2015; Lei et al., 2013) to learn local filters and encoding codebooks for better distinctiveness and compactness, respectively. Though resolved the lack of necessary properties, these shallow representations demonstrated a loss of robustness against complicated nonlinear facial appearance variations.

These traditional methods attempted to recognize faces by one- or two-layer representations and improved FR accuracy very slowly. They planned to explore each aspect of unconstrained facial variations, including illumination, pose, expression, or occlusion, separately. The advent of deep learning methods resolved the insufficiencies of traditional methods. In deep-learning-based FR approaches, multiple layers of processing units learn multiple levels of representations that correspond to different levels of abstraction. Interestingly, the higher-level abstract representations have demonstrated a strong invariance against face illumination, pose, expression, and occlusion changes, and represented facial identity with extraordinary stability. In 2014, DeepFace (Taigman et al., 2014) attained state-of-the-art accuracy on the Labeled Faces in the Wild (LFW) dataset (Huang et al., 2008). In an unconstrained condition, it competed successfully with the human performance for the first time and approached the desired accuracy by training a 9-layer network on 4 million facial images. Deep learning techniques have reformed the research horizon of FR in almost all aspects, from algorithm designs and training/test datasets to application setups and evaluation protocols.

*2.2. Distinguished Architectures of Face Recognizers*

DeepFace was the first distinguished deep architecture introduced to the FR community. It has a deep CNN architecture with several locally connected layers. Afterward, FaceNet (Schroff et al., 2015) and VGG-Face (Parkhi et al., 2015) deep-learning-based models were introduced, which were designed to train popular GoogleNet (Szegedy et al., 2015) and VGGNet (Simonyan and Zisserman, 2014) over the large-scale face datasets, respectively. These models fine-tuned the networks via a triplet loss function and implemented it on face patches created by an online triplet mining method. Later, the SphereFace (Liu et al., 2017) was proposed according to ResNet architecture (He et al., 2016), and a novel angular softmax loss learn discriminative features by an angular margin. Similar to this network, CosFace (Wang et al., 2018) and ArcFace (Deng et al., 2019) were introduced based on cosine/ angular margin-based loss, respectively. These models were designed in a way to separate learned features with a larger cosine/angular distance. Lightweight networks were then proposed to overcome the lack of GPUs' power and memory size and become applicable to many mobiles and embedded devices. Light CNN (Wu et al., 2018), with a novel max-feature-map (MFM) activation function, is a famous example of this category that results in a compact representation and reduces the computational cost.

*2.3. Standard Face Recognition Datasets*

In 2007, the LFW dataset was provided from 3K images of faces on the web under unconstrained conditions and opened a new path for other testing databases to be used in different tasks. Having sufficiently large training datasets to evaluate the effectiveness of deep FR models resulted in continually developing more complex datasets to facilitate the FR research. The early deep FR models, such as DeepFace, FaceNet, and DeepID (Sun et al., 2014), were trained on private, controlled, or small-scale training datasets, hence, not allowing the new models to compare with. To resolve this issue, CASIA-Webface (Yi et al., 2014), a collection of 0.5M images of 10K celebrities, was introduced as the first widely used public training dataset. Later, MS-Celeb-1M (Guo et al., 2016), VGGface2 (Cao et al., 2018), and Megaface (Kemelmacher-Shlizerman et al., 2016), collections of over 1M images, were introduced as a public large-scale training dataset to be used by many advanced deep learning methods.

## 3. Adversarial Attack Generation

An adversarial attack consists of finely modifying an original image with the intention of the alterations become almost imperceptible to the human eye, to fool a specific classifier. In the realm of digital attacks, this can be implemented as the addition of a minimal vector $n$ to the input image $x$, i.e. $(x + n)$, such that the deep learning model $\mathcal{F}$ predicts an incorrect output for the altered input $x + n$, which is known as an adversarial example. This way, a box-constrained optimization problem for generating the adversarial example $x'$ can generally be described as:

$$\min_{x'} \|x' - x\|_2$$
$$\text{s.t.} \ \mathcal{F}(x') = l'$$
$$\mathcal{F}(x) = l \quad (1)$$
$$l \neq l'$$
$$x' \in [0,1]$$

where $l$ and $l'$ represent the output label of $x$ and $x'$, and $\|\cdot\|_2$ denotes the distance between two image samples according to $L_2$-norm.

As represented in Fig. 2, to fool the FR model (VGG16 in this case), the input images are altered so that while the human can still forecast the correct class, the network will be confused and misled to the wrong category. Szegedy et al., (2013) were the first to demonstrate the vulnerabilities of CNN models to adversarial attacks generated by introducing a minute noise in the input image. The accuracies of GoogleNet and VGG-Face models also demonstrated to be degraded with color balance manipulation. Note that adversarial attacks' invisibility and the widespread application of deep learning algorithms can cause severe damages in real-world scenarios (Kurakin et al., 2016a). For example, if the signboard is altered in self-directed driving, adversarial examples can overly threaten the car, pedestrians, and other automobiles. Similarly, in FR applications, the failure to verify the altered input could lead to the degraded performance that can take benefit in the closed-set verification scenarios.

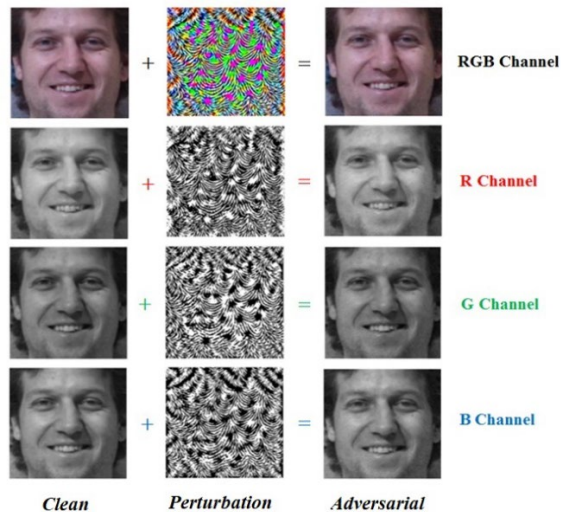

Fig. 2. Visualization of original face image (first column), adversarial noise vector of VGG-16 (second column), and altered image (last column). From top to bottom, the four rows represent the addition of adversarial noise to the original RGB image and corresponding grayscale representations of R, G, and B color *channels.* Adversarial noise is magnified by a factor of 4 to enhance visibility (Agarwal et al., 2018).

*3.1. Terms and Definitions*

This section gives a brief introduction to the standard terms related to adversarial attacks on (deep) FR models. Our definitions of words are essential to understand the technical components of the reviewed studies. The remainder of this article follows the same definitions of the terms.

*3.1.1. General Terms*

1) ***Adversarial example/image*** is an intentionally altered (e.g., by adding noise) version of a clean image to fool machine learning (ML) models, such as FR models.
2) ***Adversarial training*** is a training process that uses adversarial images along with clean images.
3) ***Adversary*** is an agent who creates an adversarial example or the example itself, depending on the case study.
4) ***Threat model*** is a model that formalizes assumptions about the attacker's goals, attack strategy, knowledge of the attacked system, and capability of employing the input data/system components concerning the target model.

*3.1.2. Specific Terms*

1) ***Dodging attack*** occurs when the attacker tries to have a face misidentified as any other arbitrary face.
2) ***Evasion attack*** tries to evade the system by altering samples during the testing phase yet not influencing the training data.
3) ***Impersonation attack*** seeks to disguise a face as a specific (authorized) face.
4) ***Poisoning attack*** takes place during the training time to contaminate the training data. In this attack, the attacker tries to poison data by inserting wisely designed samples to ultimately compromise the whole learning process.

*3.2. Adversarial Attacks Attributes*

In this section, we discuss the main attributes of adversarial example generation methods.

3.2.1. *Adversarial Capacity*

The adversarial capacity is determined by the amount of knowledge the attackers could gain about the model. Threat models in deep FR systems are classified into the following types according to the attack's capacity.

1) ***White-box attack*** assumes the complete knowledge of the target model, i.e., its parameters, architecture, training method, and even in some cases, its training data.
2) ***Black-box attack*** feeds a target model with the adversarial examples (during testing) created without knowing that model (e.g., its training procedure or its architecture or its parameters). Though the knowledge of the model is not available, the attackers can interact with such a model by utilizing the transferability of adversarial examples (Section 3.2.3).

*3.2.2. Adversarial Specificity*

Adversarial specificity is defined as the ability of the attack to allow a specific intrusion/disruption or create general mayhem. Threat models in deep FR systems could be categorized into the following types according to the attack's specificity.

1) ***Targeted attack*** deceives a model into falsely predicting a specific label for the adversarial example. In an FR or biometric system, this is achieved by impersonating distinguished people.

2) ***Non-targeted attack*** predicts the adversarial examples' labels irrelevantly, as long as the results are not the correct labels. In an FR/biometric system, this is accomplished through face dodging. Non-targeted attacks are more comfortable to implement than targeted attacks since it has more choices and space to alter the output.

*3.2.3. Adversarial Transferability*

An adversarial example's ability to continue to impact the models other than the one employed to create it is a common property called transferability. It is critical for black-box attacks where access to the target model, the training dataset, and other learning parameters may not be available. A substitute neural network model can be trained in such circumstances, and then adversarial examples can be generated against the substitute model. Due to transferability, the target model will be vulnerable to these adversarial examples. The transferability of adversarial examples could be defined from easy to hard, according to the state of having the same neural network architectures but different datasets or having different neural network architectures from the beginning (Yuan et al., 2019).

*3.2.4. Adversarial Perturbations*

Adversarial perturbation is a kind of disruption that can fool a given model on a specific image with high probability. Small perturbation is a central premise for adversarial examples. In the realm of adversarial machine learning, the goal is to minimize the norm of the smallest adversarial perturbation to make target models misclassified. Explicitly, given an input image $x$, the perturbation vector $n$ aims to alter the label of $x$, corresponding to the minimal distance from $x$ to the decision boundary of the classifier:

$$\min_{n \in \mathsf{R}^d} \|n\|_2$$
$$s.t. \ \mathsf{F}(x)\mathsf{F}(x+n) \leq 0 \tag{2}$$

where $\mathsf{R}^d$ is the dimension of the input image and perturbation vector. The perturbation could be categorized into the following types according to the scope of its implementation.

1) ***Image-specific perturbations*** can be explicitly generated according to the given input images.
2) ***Universal perturbations*** can be generated without knowing the underlying details of the given images. Note that the universality refers to the characteristic of a perturbation to have a good transferability and the ability to be applied to all input data uniformly. Although universal perturbations make it easier to create adversaries in real-world applications, most present attacks generate image-specific perturbations. It is aimed to move toward this direction and create universal perturbations that are not required to be reformed when the input samples are changed (Section 6).

*3.3. Experimental Standards*

The performance of adversarial attacks against FR systems is evaluated based on different datasets and target models. This spectrum results in complications to evaluate the adversarial attacks and quantify the robustness of FR models. Large datasets and complex models usually make the attack and defense exertions harder.

3.3.1. *Datasets*

The LFW, CASIA-WebFace, MegaFace, VGGFace2, and CelebA (Liu et al., 2015) are the most widely used image classification datasets to evaluate adversarial attacks on FR systems.

3.3.2. *Target Models*

Adversaries broadly attack several eminent deep FR models, such as DeepFace, FaceNet, VGG-Face, DeepID, SphereFace, CosFace ArcFace, OpenFace (Amos et al., 2016), dlib ("dlibC++Library," 2018) and LResNet100E-IR Face ID model ("InsightFace Model Zoo, LResNet100E-IR, ArcFace@ms1m-refine-v2.," 2018). According to these datasets and target models in the following sections, we will inspect recent studies on adversarial examples targeted FR models according to these datasets and target models.

*3.4. Pioneer Methods*

In this section, we review several pioneer methods for generating adversarial examples. Almost each one of these methods forms the basis of the real-world attacks and has the power of significantly affecting machine learning target models in practice. Descriptions provided here will show the gradual improvements of the adversarial attacks and the extent to which state-of-the-art adversarial attacks can achieve. We will focus on the main methods that attack DNNs in general and review them in chronological order to maintain discussion flow.

1) *L-BFGS*

Szegedy et al. (2013) first generated adversarial examples using an L-BFGS method. The box-constrained L-BFGS is used for approximately solving the following problem:

$$\min_{x'} \; c\|n\|_2 + \mathsf{L}(x',l) \quad \text{s.t.} \; x' \in [0,1] \tag{3}$$

where $\mathsf{L}(x',l)$ computes the classifier's loss, and a minimum $c > 0$ is approximately calculated by line-searching to satisfy the above condition. Authors showed that the above method could compute perturbations that fool neural networks when added to clean images while remains imperceptible to human eyes.

2) *Fast Gradient Sign Method (FGSM)*

Goodfellow et al. (2014) proposed a fast and straightforward method, named *Fast Gradient Sign Method (FGSM)* to compute an adversarial perturbation by solving the following problem efficiently:

$$n = \dot{o} \, sign(\nabla_x \mathsf{J}(\theta, x, l)) \tag{4}$$

where $\dot{o}$ is the perturbation magnitude, $sign(.)$ denotes the sign function, and $\nabla_x \mathsf{J}(.,.,.)$ represents the gradient of the cost function around the current value of the model parameters with respect to the $x$. The generated adversarial example $x'$ is calculated as $x' = x + n$. With the application of the *FGSM* method, adversarial examples are not computed iteratively but in a one-step gradient update along the direction of the gradient sign at each pixel. Miyato et al. (2018) proposed a closely related method and named it *Fast Gradient* $L_2$. With this method, the perturbation is computed as:

$$n = \dot{o} \frac{\nabla_x \mathsf{J}(\theta, x, l)}{\|\nabla_x \mathsf{J}(\theta, x, l)\|_2} \tag{5}$$

As it is shown, the computed gradient is normalized with its $L_2$-norm. An alternative of using the $L_\infty$-norm for normalization was proposed by Kurakin et al. (2016b) and referred to as the *Fast Gradient* $L_\infty$ method. In the literature, all of these methods are categorized as one-step methods.

3) *Basic & Least-likely Iterative Class Methods*

Kurakin et al. (2016a) extended the one-step gradient ascent idea and proposed the *Basic Iterative Method (BIM)*. The *BIM* iteratively adjusts the direction that increases the loss of the classifier by running multiple small steps. In each iteration, the values of the pixels of the image are clipped as follows:

$$x'^{(i+1)} = Clip_{\dot{o}} \left\{ x'^{(i)} + \alpha \cdot sign\left(\nabla_{x'^{(i)}} \mathsf{J}\left(\theta, x'^{(i)}, l\right)\right) \right\} \tag{6}$$

where $x'^{(i)}$ denotes the generated adversarial example at the $i^{th}$ iteration, $Clip_{\partial}\{.\}$ confines its change in each iteration, and $\alpha$ is the step size. The initialization of the *BMI* algorithm is done by setting $x'^{(0)} = x$, and its termination is controlled by the number of iterations determined by $\min(\partial + 4, 1.25\partial)$. This method is also known as the *Iterative Fast Gradient Sign Method (I-FGSM)* in the literature. Following this methodology, the *Iterative Fast Gradient Value Method (I-FGVM)* is proposed, which differs in how it uses the $\nabla_{x'^{(i)}} J$ gradient (Kurakin et al., 2016a; Rozsa et al., 2016). Specifically, the *I-FGVM* changes the input $x$ in the direction of the gradient, while the *I-FGSM* uses only the sign gradient. In each iteration of *I-FGSM*, the values of the pixels of the image are clipped as follows:

$$x'^{(i+1)} = Clip_{\partial}\left\{x'^{(i)} + \alpha \cdot \nabla_{x'^{(i)}} J\left(\theta, x'^{(i)}, l\right)\right\} \tag{7}$$

In another try, Kurakin et al. (2016a) extended *BIM* to *Iterative Least-likely Class Method (ILCM)*, similar to what they did to extend *FGSM* to its "one-step target class." They substituted the label $l$ of the image in Eq. (6) by the least likely class $ll$ predicted by the classifier and tried to maximize the cross-entropy loss.

4) *Jacobian-based Saliency Map Attack (JSMA)*

Papernot et al. (2016a) designed an adversarial attack by confining the $L_0$-norm of the perturbations. In contrast to perturbing the whole image, they planned to perturb a few pixels in the image that might induce significant changes to the output. Accordingly, they defined a saliency adversarial map, called *Jacobian-based Saliency Map Attack (JSMA)*, by which they could monitor the effect of changing each pixel of the clean image on the resulting classification. The proposed algorithm is repeated until the maximum number of allowable pixels are altered in the adversarial image so that the neural network fooling succeeded.

5) *One Pixel Attack*

J. Su et al. (2019) proposed a successful method of fooling different neural networks by only changing one pixel per image. The optimization problem becomes:

$$\begin{aligned}&\min_{x'} J\left(\theta, F\left(x'\right), l'\right)\\&s.t.\ \|n\|_0 \leq \partial_0\end{aligned} \tag{8}$$

To modify only one pixel, $\partial_0$ is set to 1, hence, making the optimization problem hard. So, the authors applied the concept of Differential Evolution (Das and Suganthan, 2010) to find the optimal solution. This technique requires the probabilistic labels predicted by the targeted model and does not necessitate any information about the network parameter values or gradients. It is implemented in a simple evolutionary strategy yet successfully fooling networks.

6) *DeepFool*

Moosavi-Dezfooli et al. (2016) proposed an iterative manner, called *DeepFool*, to find a minimal norm adversarial perturbation for a clean input image. The proposed algorithm initializes with the assumption that the input image is located in a region confined by an affine classifier's decision boundaries, and the class label of the input is initially decided. At each iteration, the image is perturbed by a small vector. It is sought to lead the resulting perturbed image to the boundaries obtained by linearly approximating the region's boundaries within which the image resides. In each iteration, the perturbations are added to the image and accumulated to compute the ultimate perturbation, which alters the input image label according to the image region's original decision boundaries. *DeepFool* has been demonstrated to provide smaller perturbations compared to *FGSM* and *JSMA* while having similar fooling ratios.

7) *Universal Adversarial Perturbations*

In contrast to their *DeepFool* method that computes image-specific perturbations, Moosavi-Dezfooli et al. (2017) proposed their newer algorithm to generate image-agnostic *Universal Adversarial Perturbations* to fool a network on any image successfully. They attemped to find a universal perturbation that satisfies the following constraint:

$$P(\mathbf{F}(x) \neq \mathbf{F}(x+n)) \geq \delta$$
$$s.t. \|n\|_p \leq \xi \tag{9}$$

where $P(.)$ denotes the probability, $\delta$ controls the fooling rate, $\|.\|_p$ refers to $L_p$-norm, and $\xi$ confines the size of universal perturbation. Accordingly, the smaller the value of $\xi$, the more imperceptible the adversarial example to human eyes. It is shown that the *Universal Adversarial Perturbations* could be generalized well across popular deep learning architectures (e.g., VGG, CaffeNet, GoogLeNet, ResNet).

8) *Carlini & Wagner Attacks (C&W)*

Carlini and Wagner, (2017) introduced a set of adversarial attacks to defeat defensive distillation. According to their study, the $L_0$-, $L_1$- and $L_2$-norms of quasi-imperceptible perturbations are restricted to fail defensive distillation for the targeted networks. It is also demonstrated that the adversarial examples generated with un-distilled networks transfer well to the distilled networks making the generated perturbations proper for black-box attacks. Regarding definition, distillation is referred to as a training procedure to transfer knowledge of a more complex network to a smaller network. This notion was initially introduced by Hinton et al. (2015). Later, Papernot et al. (2016b) introduced the variant of the procedure using the network's knowledge to improve its robustness.

## 4. Adversarial Example Generation against Face Recognition

In this section, we review adversarial examples generated against FR systems. We first explain the main attack generation methods introduced in the literature. Next, we compare different attacks according to their orientation. Finally, we repeat the comparison this time based on attributes of the adversarial capacity, specificity, transferability, and the perturbation type.

*4.1. Methods*

In this section, we review the main adversarial example generation methods against FR models. We review different studies in which they will be compared in succeeding sections to maintain the discussion flow.

1) *Image-level Grid-based Occlusion*

Distortions that are not specific to faces and can be applied to any object image are categorized as image-level distortions. Goswami et al. (2018) introduced an image-level distortion called *Grid-based Occlusion*. In this approach, a number of points $P = \{p_1, p_2, \ldots, p_n\}$ are selected along the image's upper ($y = 0$) and left ($x = 0$) boundaries according to a parameter $\rho_{grids}$, where grids refer to *Grid-based Occlusion*. The $\rho_{grids}$ parameter determines the number of grids utilized to alter the given image with higher values to result in a denser grid, i.e., more grid lines. For each point $p_i = (x_i, y_i)$, a point on the opposite boundary of the image, $p_i' = (x_i', y_i')$, is selected, with the condition if $y_i = 0$ then $y_i' = H$, and if $x_i = 0$ then $x_i' = W$, where $W \times H$ is the input image's size. Once a set of pair points $P$ and $P'$ selected, one-pixel wide lines are created to link each pair. Finally, the pixels placed on these lines set to 0 grayscale value.

2) *Image-level Most Significant Bit-based Noise (xMSB) Distortion*

Image-level most significant bit-based noise is another image-level distortion introduced by Goswami et al. (2018). In this approach, three sets of pixels $\mathsf{X}_1$, $\mathsf{X}_2$, $\mathsf{X}_3$ are selected stochastically from the image such that $|\mathsf{X}_i| = \varnothing_i \times W \times H$. Here $W \times H$ is the input image size, and the parameter $\varnothing_i$ represents the fraction of pixels where the $i^{th}$ most significant bit is flipped. Accordingly, the higher the value of $\varnothing_i$, the more pixels are distorted in the $i^{th}$ most significant bit. For each $P_j \in X_1, \forall i \in [1,3]$, the following operation is pursued:

$$P_{kj} = P_{kj} \oplus 1 \qquad (10)$$

where $P_{kj}$ represents the $k^{th}$ most significant bit of the $j^{th}$ pixel in the set and $\oplus$ denotes the bitwise XOR operation. Also, it should be noted that the sets $X_1$ may overlap; hence, the total number of pixels influenced by the noise is less than or equal to $|X_1 + X_2 + X_3|$, depending on the stochastic selection.

3) *Face-level Distortion*

Besides image-level distortion, Goswami et al. (2018) also introduced face-level distortions. This type of distortion expressly necessitates face-specific information, e.g., location of facial landmarks. As a result, this approach is typically applied after performing automatic face and facial landmark detection. Once facial landmarks are detected, they are utilized along with their boundaries to perform the masking step. To obscure the eye region, a singular blocking band is drawn on the face image as follows:

$$I\{x,y\} = 0, \forall x \in [0, W], y \in \left[ y_e - \frac{d_{eye}}{\psi}, y_e + \frac{d_{eye}}{\psi} \right]$$

(11)

where $y_e = \left( \frac{y_{le} + y_{re}}{2} \right)$, and $(x_{le}, y_{le})$ and $(x_{re}, y_{re})$ are positions of left eye center and right eye center, respectively. The $d_{eye}$ is the inter-eye distance and calculated as $x_{re} - x_{le}$, and $\psi$ is the parameter that determines the occlusion band's width. The *Eye Region Occlusion* (*ERO*) process could be implemented to obscure forehead and brow in a similar trend using the facial landmarks on the forehead and brow regions as a mask. It could also be implemented to occlude beard region utilizing the outer facial landmarks and nose and mouth coordinates to create the mask as combinations of individually occluded regions.

4) *Evolutionary Attack*

Dong et al. (2019) proposed *Evolutionary Attack* method, based on *(1+1)-CMA-ES* (Igel et al., 2006), which is a useful and straightforward variant of the *Covariance Matrix Adaptation Evolution Strategy* (*CMA-ES*) (Hansen and Ostermeier, 2001). In each update iteration of the *(1+1)-CMA-ES*, a new offspring (candidate solution) is generated from its parent (current solution) by adding random noise, the objective of these two solutions is evaluated, and the better one is selected for the next iteration. This method is capable of solving the black-box optimization problem of:

$$\min_{x'} L(x') = \|x' - x\|_2 + \delta(C(F(x')) = 1) \qquad (12)$$

where $C(.)$ is an adversarial criterion that takes 1 if the attack requirement is satisfied and 0 otherwise, and $\delta(a)$ is 0 if $a$ is true, and $+\infty$, otherwise. However, the authors did not apply the *(1+1)-CMA-ES* to optimize Eq. (12) due to the high dimension of $x'$. To accelerate this algorithm, they proposed an appropriate distribution to sample the random noise in each iteration, which can model the local geometry of the search directions. They sampled a random noise from a biased Gaussian distribution to minimize the sampled adversarial image's distance from the original image. This added bias term is a critical hyper-parameter controlling the strength of going towards the original image. The authors also proposed techniques to reduce the search space's dimension by considering the particular characteristics of this problem. They sampled random noise in a lower-dimensional space $R^m$ with $m < d$, where $d$ is the dimension of input space. They then adopted an upscaling operator, precisely, the bilinear interpolation method, to project noise vector to the original space. Consequently, the input image dimension is preserved, and the dimension of search space is reduced.

5) *Feature Fast & Iterative Attack Methods*

Given a face pair and a deep face model, Zhong and Deng (2019) proposed feature-level attacks to compare the face pair via calculating the distance between their normalized deep representations. These representations are similar to the embedding features, except that they are normalized and extracted from the deep face model. To discover the vulnerability of deep face models, the authors proposed to add perturbation on one of the face images to generate adversarial examples and deceive the face model. According to their notion, a positive and negative face pair is defined, for which the corresponding output labels are the same and different, respectively. Denoting the face pair by $\{x^1, x^2\}$ and adversarial example by $x' = x^1 + n$, for a positive face pair, $l^1 = l^2$ and the optimized objective and loss function are formulated as:

$$\begin{aligned} n &= \underset{n}{\operatorname{argmax}} \left\| F(x^1 + n) - F(x^2) \right\|_2, \quad \|n\|_\infty < \varepsilon \\ J(x^1 + n, x^2) &= \left\| F(x^1 + n) - F(x^2) \right\|_2 \end{aligned} \tag{13}$$

while for negative face pair $\{x^1, x^2\}$, $l^1 \neq l^2$, and the optimized objective and loss function is formulated as:

$$\begin{aligned} n &= \underset{n}{\operatorname{argmax}} \left\| F(x^1 + n) - F(x^2) \right\|_2, \quad \|n\|_\infty < \varepsilon \\ J(x^1 + n, x^2) &= -\left\| F(x^1 + n) - F(x^2) \right\|_2 \end{aligned} \tag{14}$$

where $F(x^i)$ denotes deep representations after normalization and $\varepsilon$ limits the maximum deviation of the perturbation. Forming adversarial perturbation based on the loss functions of Eq. (13) and Eq. (14) is called *Feature Fast Attack Method (FFM)* and defined as:

$$x^1 + n = \mathbf{G}_{x^1, \varepsilon}\left(x^1 + sign\left(\nabla_{x^1} J(x^1, x^2)\right)\right) \tag{15}$$

Considering an iterative way, the authors proposed the *Feature Iterative Attack Method (FIM)* as:

$$\begin{aligned} n_0 &= 0 \\ g_{N+1} &= \nabla_{x^1 + n_N} J(x^1 + n_N, x^2) \\ x^1 + n_{N+1} &= \left(x^1 + n_N, sign(g_{N+1})\right) \end{aligned} \tag{16}$$

where $\mathbf{G}_{x, \varepsilon}(x') = \min(255, x + \varepsilon, \max(0, x - \varepsilon, x'))$; the iteration can be chosen heuristically $\min(\varepsilon + 4, 1.25\varepsilon)$.

6) *Eyeglass Accessory Printing*

Sharif et al. (2016) proposed a physically realizable attack for impersonation or dodging in a digital environment. To enable physical realizability, the first step involved implementing the attacks purely with facial accessories (specifically, eyeglass frames) via 3d- or even 2d-printing technologies. In particular, they used a specific readily available digital model of eyeglass frames and utilized a commodity inkjet printer (Epson XP-830) to print the front plane of the eyeglass frames on glossy paper, which are affixed to actual eyeglass frames, subsequently. After alignment, the frames occupy about 6.5% of the 224 × 224 face image pixels, implying that the attacks perturb at most 6.5% of the pixels in the image. To find the color of the frames necessary to achieve impersonation or dodging, their color is initialized to a solid color (e.g., yellow), and the frames are rendered onto the image of the subject. Their color is updated iteratively through the gradient descent process to craft adversarial perturbations tolerant to slight natural movements when physically wearing the frames.

The second step involved tweaking the mathematical formulation of the attacker's objective to focus on adversarial perturbations that both robust to small changes in viewing condition and smooth as expected from natural images. To find perturbations independent of the exact imaging conditions, aiming to enhance the generality of the perturbations, the authors looked for perturbations that can cause any image in a set of inputs to be misclassified. To this end, an attacker collects a set of images, $X$, and finds a single perturbation that optimizes her objective for every image $x \in X$. For impersonation, this is formalized as the following optimization problem (dodging is analogous):

$$\underset{n}{\text{argmin}} \sum_{x \in X} softmaxloss(\mathbf{F}(x+n), l) \tag{17}$$

where $n$ denotes the perturbation. To preserve the smoothness of perturbations, the optimization is updated to account for minimizing total variation (TV) (Mahendran and Vedaldi, 2015), which is defined as:

$$TV(n) = \sum_{i,j} \left( \left( n_{i,j} - n_{i+1,j} \right)^2 + \left( n_{i,j} - n_{i,j+1} \right)^2 \right)^{1/2} \tag{18}$$

where $n_{i,i}$ denotes a pixel in $n$ at coordinate $(i, j)$. $TV(n)$ is low when the values of adjacent pixels are close to each other (i.e., the perturbation is smooth), and high otherwise. Therefore, by minimizing $TV(n)$, the smoothness of the perturbed image hence the physical realizability is improved.

7) *Visible Light-based Attack (VLA)*

Shen et al. (2019) introduced a *Visible Light-based Attack* (*VLA*) against FR systems, where visible light-based adversarial perturbations are crafted and projected on human faces. For each adversarial example, the authors proposed generating a perturbation frame and a concealing frame, projecting the two frames to the user's face. The perturbation frame contains information on how to change the input user's facial features to the features of a targeted or non-targeted user, while the concealing frame aims to hide the perturbations in the perturbation frame from being observed by human eyes.

Regarding the perturbation frames generation, this method enlarges the pixel-level image modifications into region-level to avoid probable perturbation loss in physical scenarios. Accordingly, the perturbation frame is divided into exclusive ranges based on the similarity of containing color values. A MeanShift clustering does the division over all colors, where nearby similar colors are divided into the same regions, and each group of nearby pixels with the same color in the image is regarded as one perturbation region. Then, in the second step, a region filtering strategy is utilized to ensure that the camera can successfully capture all projected details in a perturbation frame, and small color regions would not get lost in the images captured in physical scenarios. Denoting $n = x' - x$ as the perturbation frame, a clustering and filtering result of $n$ is denoted by $C_{x,x'}$ and defined as follows:

$$C_{x,x'} = \{G_i(p), R_i \mid 0 \leq i \leq m\} \tag{19}$$

where $G_i(p)$ indicates whether the color of a pixel $p$ should be set as $R_i$, and $m$ is the total number of color regions. For each pixel $p$. in the image $C_{x,x'}$, $G_i(p)$ is 1 if $p$ lies within $R_i$, and 0, otherwise. The generation function $\mathsf{H}(\cdot)$ is defined next to transform the clustering result $C_{x,x'}$ into a perturbation frame $n$, as shown in Eq. (20):

$$n = \mathsf{H}(C_{x,x'}) = \left[ R_i \text{ if } G_i(p) = 1 \right] \tag{20}$$

To hide the perturbation frames from human eyes, concealing frames are generated according to the effect of *Persistence of Vision* (*POV*) (Zhang et al., 2015). According to *POV*, two different colors that swap frequently cause the human brain not directly process these changes at the exact moment they occur, making the human eyes perceive a new color as a fusion of those colors. Based on this knowledge, by projecting the perturbation frame and the concealing frame alternately, i.e., displaying the corresponding two colors of generated images interchangeably, it can be difficult for human eyes to feel the perturbation frame, and a fusion of these colors will be perceived as a base/background color of the image.

8) *AdvHat attack*

Komkov and Petiushko (2019) proposed a reproducible adversarial attack generation method, called *AdvHat*. They printed a rectangular paper sticker on a standard color printer and put it on the hat with an off-plane transformations algorithm. The proposed algorithm split into two steps: (1) off-plane bending of the sticker, which is simulated as a parabolic transformation in the 3D space to map each point of the sticker to the new point on the parabolic cylinder, and (2) pitch rotation of the sticker, which is stimulated by the application of a 3D affine transformation to the obtained new points. The authors projected the resulted sticker on the high-quality face image with small perturbations in the projection parameters. They transformed the new face image into the standard template of ArcFace input to pass it to the optimization step. Regarding the optimization step, the sum of two parameters (TV loss and cosine similarity between two embeddings) is minimized as follows to achieve the gradient signs used to modify the sticker image:

$$L_T(x', a) = L_{sim}(x', a) + \lambda \cdot TV(patch) \tag{21}$$

where $L_T$ is the total loss, *patch* denotes the sticker, $x'$ is a photo with the applied patch, and $\lambda$ is a weight for TV loss, which is assumed to be $1e-4$ in this work. Here, $L_{sim}$ is cosine similarity between two embeddings and defined as follows:

$$L_{sim}(x', a) = \cos(e_{x'}, e_a) \tag{22}$$

where $e_{x'}$ is obtained embeddings of the face image of the attacker and $e_a$ refers to the embedding of the desired person's face image calculated by ArcFace.

9) *Penalized Fast Gradient Value Method (P-FGVM)*

Chatzikyriakidis et al. (2019) introduced a *Penalized Fast Gradient Value Method* (*P-FGVM*) adversarial attack technique, which runs on the image spatial domain and generates adversarial de-identified facial images like the original ones. This technique is inspired by the *I-FGVM*, with a minor exception of combining an adversarial loss and a 'realism' loss term in its gradient descent update equations. In this method, a targeted adversarial example $x'$ is generated through the following gradient descent update equations:

$$x'^{(i+1)} = Clip_\delta \left\{ x'^{(i)} + \alpha \cdot \left( \nabla_{x'^{(i)}} J(\theta, x'^{(i)}, l) + \lambda(x'^{(i)} - x) \right) \right\} \tag{23}$$

where $\lambda$ is a weight coefficient and $(x'^{(i)} - x)$ is the realism loss term.

10) *Face Friend-safe Attack*

Kwon et al. (2019) proposed the *Face Friend-safe* adversarial example generation method, which generates adversarial examples that are misrecognized by an enemy FR system, nonetheless, appropriately recognized by a friend FR system with the least distortion. The proposed method consists of a transformer, a friend classifier $M_{friend}$, and an enemy classifier $M_{enemy}$, to generate adversarial face images. Given the pre-trained $M_{friend}$ and $M_{enemy}$ and the original input $x \in X$, the optimization problem of generating the adversarial face example $x'$ is as follows:

$$\operatorname*{argmin}_{x'} L(x, x')$$
$$\text{s.t. } g^{friend}(x') = l \text{ and } g^{enemy}(x') \neq l \tag{24}$$

where $g^{friend}(x)$ and $g^{enemy}(x)$ denote the operation functions of a friend classifier $M_{friend}$ and enemy classifier $M_{enemy}$, respectively. $L(.)$ is the distance measured between the face original sample $x$ and face transformed example $x'$. The transformer generates adversarial face example $x'$, taking the original sample $x$ and its corresponding output label. The classification loss of $x'$ by $M_{friend}$ and $M_{enemy}$ are returned to the transformer, which then calculates the total loss, $L_T$, and repeats the above procedure to generate an adversarial face example $x'$ while minimizing $L_T$. This total loss is defined as follows:

$$L_T = L_{friend} + L_{enemy} + L_{distortion} \tag{25}$$

where $L_{friend}$ is the classification loss function of $M_{friend}$, $L_{enemy}$ is the classification loss function of $M_{enemy}$, and $L_{distortion}$ is the distortion of the transformed example, and defined as the distance between $x$ and $x'$.

11) *Fast Landmark Manipulation (FLM) Method*

Dabouei et al. (2019) proposed a fast landmark manipulation approach to craft adversarial faces. They proposed to generate adversarial examples by spatially transforming original images. Using a landmark detector function $\Phi$, that maps the face image $x$ to a set of $k$ 2D-landmark locations $P = \{p_1, \ldots, p_k\}$, $p_i = (u_i, v_i)$, it is assumed that $p_i' = (u_i', v_i')$ is the transformed version of $p_i$, and defines the $i^{th}$ landmark location in the corresponding adversarial image $x'$. To manipulate the face image based on $P$, a per-landmark flow (displacement) $f$ is defined to produce the location of the corresponding adversarial landmarks. Accordingly, the adversarial landmark $p_i'$ can be obtained from the original landmark $p_i$ and optimized particular displacement vector $f_i = (\Delta u_i, \Delta v_i)$ as follows:

$$\begin{aligned} p_i' &= p_i + f_i \\ (u_i', v_i') &= (u_i + \Delta u_i, v_i + \Delta v_i) \end{aligned} \tag{26}$$

In contrast with the reference work (Xiao et al., 2018), which fulfills this purpose by defining field $f$ for all pixel locations in the input image, (Dabouei et al., 2019) defined it only for $k$ landmarks, which is notably small compared to the number of pixels in the input image, especially when incorporated in real applications like FR problems. This limited number of control points also reduces the distortion introduced by the spatial transformation. Using the transformation $T$, the benign face image spatially transformed into an adversarial face image as follows:

$$x' = T(P, P', x) \tag{27}$$

where $P'$ refers to target control points. Incorporating the softmax cost as the measure for the correct classification, authors defined the total loss for generating adversarial faces as:

$$\mathsf{L}\left(P, P', x, l\right) = softmaxloss\left(\mathbf{F}\left(T(P, P', x)\right), l\right) - \lambda_{flow} L_{flow}\left(P' - P\right) \qquad (28)$$

where $\lambda_{flow}$ is a positive coefficient used to control the magnitude of displacement, and $L_{flow}$ is a term incorporated for bounding the displacement field. This way, the landmark displacement field $f$ is found iteratively using the gradient direction of the prediction and called the *FLM* method. Authors also extended this approach proposing the *Grouped Fast Landmark Manipulation* (*GFLM*) Method, which semantically groups landmarks and manipulates the group properties instead of perturbing each landmark. This idea was formed to resolve severe distortion of the adversarial faces generated by *FLM* and preserve the created images' whole structure.

*4.2. Comparison of Different Adversaries on Orientation*

A general taxonomy of existing adversarial example generation techniques against FR systems considering the adversaries' orientation is depicted in Fig. 3. These techniques are mainly classified into four categories, namely, (1) CNN models-oriented; (2) physical attacks-oriented; (3) de-identification-oriented; and (4) geometry-oriented. The remainder of this section is structured according to this classification.

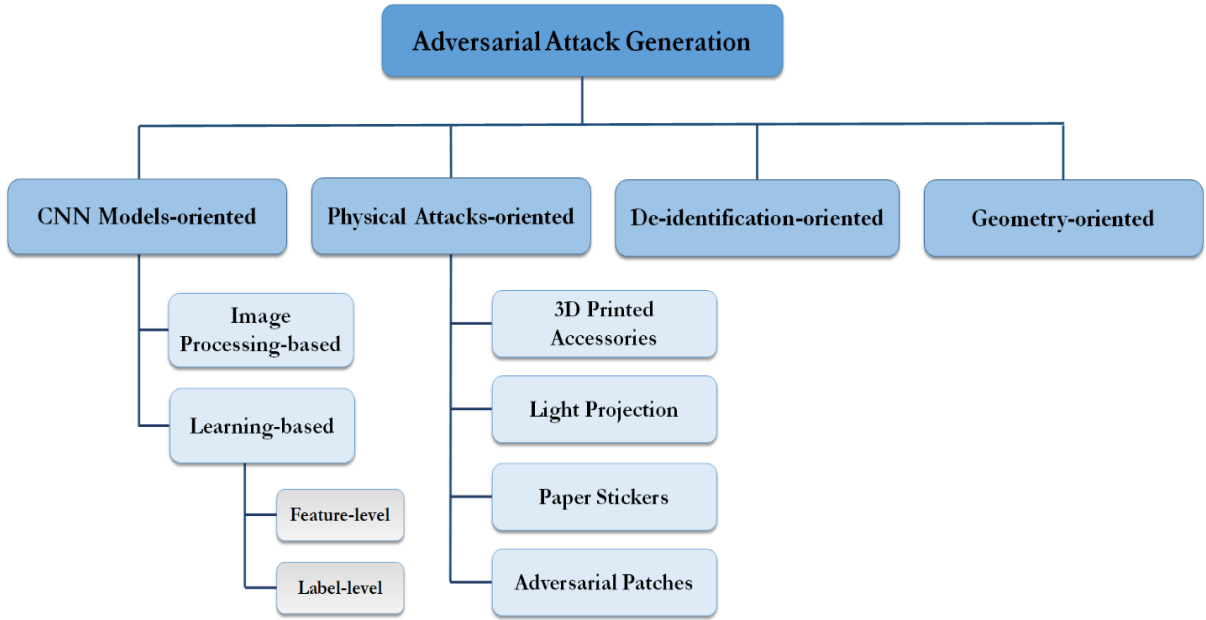

Fig. 3. The broad categorization of adversarial attack generation methods aimed to deceive the FR systems.

*4.2.1. CNN Models-oriented*

As stated earlier, the deep learning paradigm has seen a remarkable propagation in FR mission. Several models are deep CNN-based architectures with many hidden layers and millions of parameters designed to achieve very high accuracies when tested on different databases. While astonishing progress in such models' reported efficiencies improves, they are shown to be susceptible to adversarial attacks. Realizing this, many researchers have started to design approaches to exploit the weaknesses of such algorithms to investigate their robustness and revealing their singularities.

Goswami et al. (2018) considered the vulnerability of several deep CNN-based FR algorithms in the presence of image processing-based distortions at (1) image-level and (2) face-level. They confirmed that attacks to systems do not need to be sophisticated learning-based. Instead, a random noise or even horizontal and vertical black grid lines drawn in the face image can severely reduce the face verification accuracies. Examples of this effort are depicted in Fig. 4.

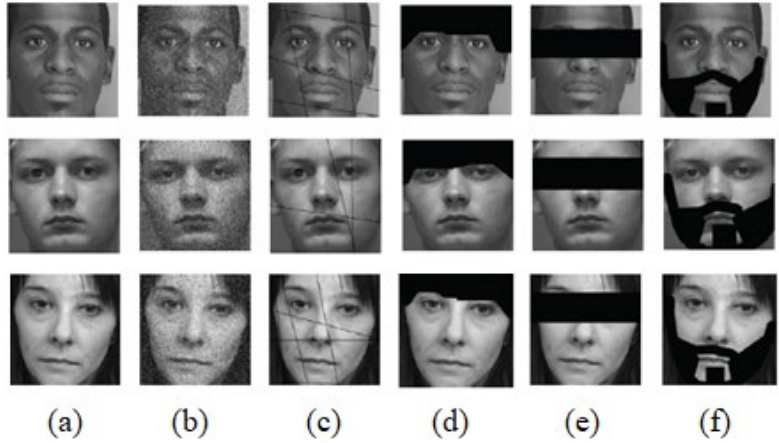

Fig. 4. Clean input images (a) modified by image processing-based distortions of *xMSB* (b), *Grid-based Occlusion* (c), *Forehead and Brow Occlusion* (*FHBO*) (d), *Eye Region Occlusion* (*ERO*) (e), and *Bread-like Occlusion* (f) (Goswami et al., 2018).

Dong et al. (2019) proposed the *(1+1)-CMA-ES evolutionary attack* algorithm to evaluate the robustness of multiple advanced FR models, including SphereFace, CosFace, and ArcFace, in a decision-based attack setting. On the LFW and MegaFace datasets, the performance of the evolutionary attack method compared with the boundary attack method (Brendel et al., 2017), optimization-based practice (Cheng et al., 2018), and an extension of NES in the label-only setting (NES- LO) (Ilyas et al., 2018). Experiments showed that against all FR models, the proposed method could converge much faster and achieve smaller distortions compared with other methods consistently.

Zhong and Deng (2020) defined *Dropout Face Attacking Networks* (*DFANet*) technique to explore the vulnerability of deep CNNs against feature-level adversarial examples. They incorporated dropout in the convolutional layers in the iterative steps of the adversarial generation process to improve the transferability of adversarial examples. Specifically, for a face model composed of convolutional layers, given the output of the $i^{th}$ convolutional layer, they proposed to generate a mask, each element of which is independently sampled from a Bernoulli distribution. this mask is then utilized to modify the output of the $i^{th}$ convolutional layer via Hadamard product of those. Authors proposed to apply this method to the generation of *FIM* and combined it with transferability enhancement methods (Dong et al., 2018; Liu et al., 2016; Xie et al., 2019). Applying their practice on the LFW dataset, they generated a new set of adversarial face pairs to attack commercial APIs of Amazon ("AmazonâA Zs Rekognition Tool," n.d.), Microsoft ("Microsoft Azure," n.d.), Baidu ("Baidu Cloud Vision Api," n.d.) and Face++ ("Face++ Research Toolkit," n.d.). They made this TALFW database available to the public for future investigations.

Recently, a new Python-based toolbox, termed Advbox, is proposed to generate adversarial examples (Goodman et al., 2020). With Advbox, it is possible to fool neural networks in PaddlePaddle, PyTorch, Caffe2, MxNet, Keras, and TensorFlow, with the additional capability to benchmark the robustness of ML models. Compared to previous works, this platform supports actual attack scenarios, such as FR attacks.

*4.2.2. Physical Attacks-oriented*

Intruders to facial biometric systems often encountered with two kinds of challenges: (1) they do not have precise control over the FR systems' (digital) input; instead, they may be able to control their physical appearance, and (2) they might be easily observed by traditional means like the police, when manipulating their appearances to evade recognition, e.g., with an excessive amount of makeup. In light of such challenges, a new class of adversarial attacks has emerged based on the attackers' physical state.

Sharif et al. (2016) developed the *Eyeglass Accessory Printing* method to generate a physically realizable yet inconspicuous class of attacks. Evaluating their method, they were able to evade recognition by changing the test inputs. In (Sharif et al., 2019), the authors defined generative adversarial nets (GANs) to attack VGG-Face and OpenFace models on both digital and physical levels of evasion purposes. The FR algorithms were targeted on the digital-level by traditional attacks, such as Szegedy's L-BFGS method (Szegedy et al., 2013), and deceived on the physical-level by requesting individuals to wear their 3D printed sunglasses frames. Fig. 5 illustrates an impersonation attack generation by wearing such an accessory.

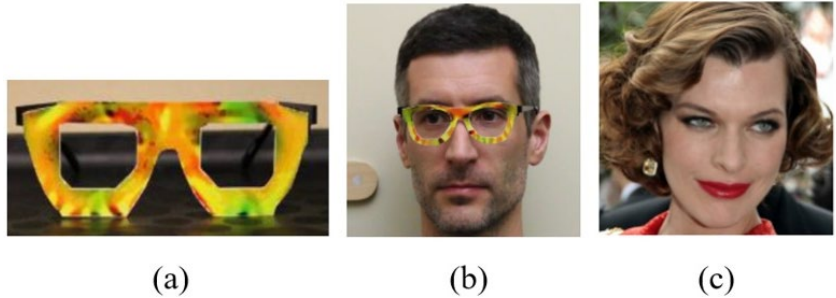

Fig. 5. The eyeglass frames (a) were used by Lujo Bauer (b) to impersonate Milla Jovovich (c) (Sharif et al., 2016).

Zhou et al. (2018) designed a cap, with some penny-size lit Infrared LEDs on the peak, to generate inconspicuous physical adversarial attacks via Infrared dot direction on the carrier's face. The loss in this work is optimized by adjusting light spots in line with the model on the attacker's photo. The attacker could then evade detection by adjusting the positions, sizes, and strengths of the dots. Using the LFW dataset, the proposed technique's effectiveness was examined on the FaceNet model, demonstrating that a single attacker could effectively target a considerable number of people.

Motivated by the differences in image-forming principles between cameras and human eyes, Shen et al. (2019) proposed the *VLA* attack against FR models. They conducted extensive experiments on the LFW dataset and against FaceNet, SphereFace, and dlib models. As compared with the *FGSM*, the proposed approach was demonstrated to achieve significantly higher success rates. Further experiments also revealed the inconspicuousness and robustness of the adversarial examples crafted by *VLA* in physical scenarios. In a similar study, Nguyen et al. (2020) studied the feasibility of directing real-time physical attacks on FR systems by adversarial light projections using a web camera and a projector. In this approach, the authors captured the adversary's facial image with a camera and used one or more target images to (1) adjust the camera-projector setup according to the attack environment and (2) create a digital adversarial pattern. The digital pattern is then projected onto the adversary's face in the physical domain with a projector to evade recognition. Although this work's objectives are identical to the infrared-based adversarial attacks (Zhou et al., 2018), this work does not necessitate creating a wearable artifact; thus, it offers a more comfortable alternative setup to direct physical attacks on FR models. Experimental results on FaceNet, SphereFace, and one commercial FR system demonstrated such models' vulnerability to light projection attacks.

Another study (Komkov and Petiushko, 2019) proposed to target the public Face ID model LResNet100E-IR, ArcFace@ms1m-refine-v2, by *AdvHat* attack generation method. On the CASIA-WebFace dataset, experimental results verified that such an approach could easily confuse the LResNet100E-IR Face ID model. Similarly, Pautov et al. (2019) examined the security of the same recognition system and proposed to print, add (as face attributes) and photograph adversarial patches; the snapshot of an individual with such attributes is then delivered to the classifier to alter the correctly recognized class to the desired one. In this work, patches were either various parts of the attacker's face, like nose or forehead, or some wearable accessories such as eyeglasses. On the CASIA-WebFace dataset and photos of the first and second authors of this work, experiments showed that such a simple attacking technique could deceive the FR system in the digital and physical worlds. In other words, the authors demonstrated that it is possible to attack ArcFace in the real world by the application of adversarial stickers on eyeglasses or forehead.

*4.2.3. De-identification-oriented*

Since the face as a biometric tool has achieved high acceptance, much effort has been made to develop its security. In return, smart adversaries aim to deny service to authentic users or let impostors evade the FR system. Considering this fact, researchers focused on the security aspect of face authentication systems.

Garofalo et al. (2018) deployed a poisoning attack against an authentication system based on the OpenFace recognition framework. They implemented the attack against the underlying *support vector machine* (*SVM*) model to classify face templates extracted by the FaceNet model. Within their evaluation framework, attacks successfully triggered remarkable authentication errors.

Chatzikyriakidis et al. (2019) proposed to utilize adversarial examples in cases of face de-identification. They introduced the *P-FGVM* adversarial attack technique and evaluated it on two CNN-based face classifiers: (1) a simple architecture model and (2) a fine-tuned model with transfer learning, based on the pre-trained VGG-Face CNN descriptor, using the VGG-16 architecture (Simonyan and Zisserman, 2014). Comparing with the baseline *I-FGVM*, against the face classifiers described above and on a subset of the CelebA dataset, the authors demonstrated that the *P-FGVM* method both protects privacy and preserves visual facial image quality more efficiently. Examples of implementing this method to generate adversarial images are shown in Fig. 6.

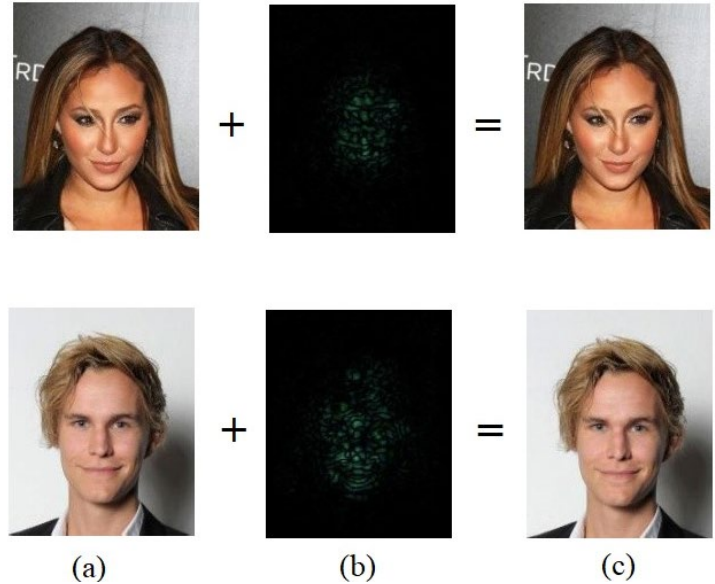

Fig. 6. Clean facial images (a) modified by adversarial perturbation (b) to generate de-identified facial images (c) via adversarial attack method *P-FGVM* (Chatzikyriakidis et al., 2019). The absolute value of perturbation is amplified by 10x.

Lately, Kwon et al. (2019) proposed the *Face Friend-safe* adversarial example generation method. Considering the FaceNet recognition system as the target model, the authors trained their method on VGGFace2 and tested it on the LFW dataset. They evaluated the efficiency of the proposed method by measuring the attack success rate of the enemy classifier, the accuracy of the friend classifier, and the average distortion, and demonstrated that the objectives of this work were achieved successfully.

*4.2.4. Geometry-oriented*

Prevalent intensity-based adversarial attack methods, which manipulate the intensity of input images directly, are computationally cheap but sensitive to spatial transformations. A small rotation, translation, or scale variation in the input image could result in a drastic change in similarity in these methods. Due to this limitation, a new class of attacks was initiated to generate geometry-based adversarial examples.

Dabouei et al. (2019) proposed the *FLM* method to craft adversarial faces almost 200 times quicker than traditional geometric attacks. They further introduced *GFLM* as the extended version of the fast geometric perturbation generation algorithm. Training FaceNet model on VGGFace2 and CASIA-WebFace datasets and evaluating its performance on CASIA-WebFace dataset, experiments revealed that both the *FLM* and *GFLM* could generate powerful adversarial face images that fool the classifier significantly. Fig. 7 demonstrates an overview of the proposed fast geometry-based adversarial attack (Dabouei et al., 2019).

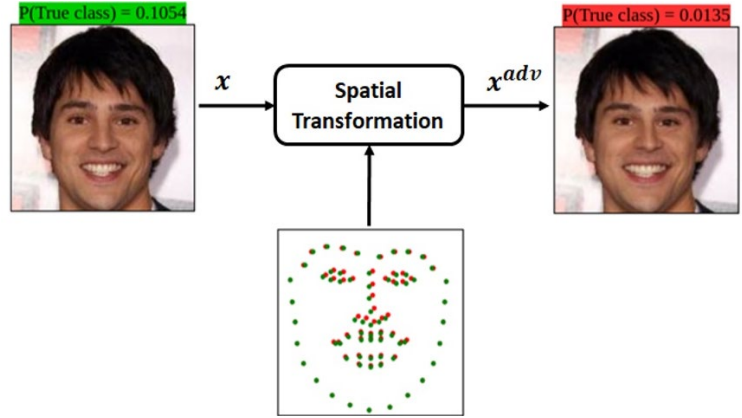

Fig. 7. Fast landmark manipulation method application to produce adversarial landmark locations, with which the ground truth image spatially transformed to a natural adversarial image. As shown in green and red colors, respectively, the ground truth image is correctly classified, while the adversarial image is misclassified to a wrong class (Dabouei et al., 2019).

Song et al. (2018) focused on attacks that mislead the FR networks to detect someone as a target person, not misclassify inconspicuously. They introduced an *attentional adversarial attack generative network* ($A^3GN$) to generate adversarial examples similar to the original images while having the same feature representation as the target face. To capture the target person's semantic information, they appended a conditional variational autoencoder and attention modules to learn the instance-level correspondences between faces. To examine the proposed method, training was accomplished on CASIA-WebFace, and evaluation was fulfilled on LFW datasets. Comparing with *stAdv* (Xiao et al., 2018) and *GFLM*, this approach achieved a satisfactory attack success rate. Overall, the authors demonstrated the excellent performance of $A^3GN$ by a set of evaluation criteria in physical likeness, similarity score, and accuracy of recognition on different target faces.

Utilizing GANs, Deb et al. (2019) crafted natural face images with a barely distinguishable difference from target face images. They proposed the *AdvFaces* adversarial face synthesis method to craft minimal perturbations in the prominent facial regions. This method comprises a generator, a discriminator, and a face matcher to automatically generate an adversarial mask added to the image to obtain an adversarial face image. Training *AdvFaces* on CASIA-WebFace and testing it on LFW, adversarial faces generated by this approach could evade several new face matching techniques and capable of achieving remarkable attack success rates. Table 1 presents a general overview of different adversarial example generation approaches regarding their orientation.

*4.3. Comparison of Different Adversaries on Attributes*

This section compares different adversarial example generation techniques in terms of attack attributes of capacity, specificity, transferability, and the kind of employed perturbation.

*4.3.1. The Capacity*

Table II summarizes two primary attribute information, i.e., the capacity and the specificity of attack methods. Regarding the capacity attribute, we find that most of the attack generation techniques are white-box attacks. In the scenario of black-box attacks, focusing on CNN model orientation, Dong et al. (2019) considered a black-box decision-based attack setting and demonstrated that their approach could converge fast and fool the target model with fine distortions. Zhong and Deng (2020) designed operative black-box adversarial attacks against commercial APIs and took a further step exploring the transferability of feature-level adversarial examples against deep CNN-based FR models (Section 4.3.3). Goodman et al. (2020) proposed the Advbox toolbox, which showed its ability to support black-box attacks against FR systems. Regarding physical attacks-orientation, Shen et al. (2019) proposed their *VLA* against lack-box FR systems, and Nguyen et al. (2020) focused on real-time light projection-based attacks considering both white- and black-box attack settings. In geometry-oriented attacks, Deb et al. (2019) demonstrated that faces generated by *AdvFaces* adversarial face synthesis method could evade several black-box contemporary face matching techniques while achieving unprecedented attack success rates.

Table 1. Comparison of different adversarial attack generation algorithms on the orientation

| Representative study | Attacks orientation | Description |
| --- | --- | --- |
| (Goswami et al., 2018) | CNN models | Adversarial image creation with distortions at image-level and face-level |
| (Dong et al., 2019) | CNN models | Decision-based attack generation with a *(1+1)-CMA-ES-based evolutionary algorithm* |
| (Zhong and Deng, 2020) | CNN models | Feature-level transferability enhancement by dropout-based *DFANet* method |
| (Goodman et al., 2020) | CNN models | Advbox toolbox |
| (Sharif et al., 2019, 2016) | Physical | Evasion attacks on digital-level with traditional L-BFGS method and physical-level with 3D printed sunglasses frames |
| (Zhou et al., 2018) | Physical | Physical adversarial examples creation with infrared LEDs attached to a cap |
| (Shen et al., 2019) | Physical | *VLA* in the physical world |
| (Nguyen et al., 2020) | Physical | Real-time light projection-based physical adversarial attacks |
| (Komkov and Petiushko, 2019) | Physical | Reproducible transferable attack on LResNet100E-IR Face ID system through projecting a paper sticker on the hat |
| (Pautov et al., 2019) | Physical | Adversarial attack on LResNet100E-IR Face ID system by printing, adding, and photographing adversarial patches of nose, forehead, and eyeglasses of the attacker |
| (Garofalo et al., 2018) | De-identification | Poisoning attack on an authenticator, based on OpenFace framework extended with an SVM classifier |
| (Chatzikyriakidis et al., 2019) | De-identification | De-identified facial images generation with *P-FGVM* adversarial attack technique |
| (Kwon et al., 2019) | De-identification | Face friend-safe adversarial examples generation |
| (Dabouei et al., 2019) | Geometric | Geometrically face transformation via fast landmark manipulation |
| (Song et al., 2018) | Geometric | Attentional adversarial attack generative network, $A^3GN$, to generate adversarial examples not misclassify inconspicuously |
| (Deb et al., 2019) | Geometric | Model-agnostic and transferable adversarial face generation via adversarial face synthesis method, *AdvFaces*, through minimal perturbations in salient facial regions |

Table 2. Comparison of different adversarial attacks on capacity and specificity attributes

| Representative study | Adversarial capacity | Adversarial Specificity |
| --- | --- | --- |
| (Goswami et al., 2018) | None | None |
| (Dong et al., 2019) | Black-box | Both |
| (Zhong and Deng, 2020) | Black-box | Targeted |
| (Goodman et al., 2020) | Both | Both |
| (Sharif et al., 2019, 2016) | White-box | Both |
| (Zhou et al., 2018) | White-box | Both |
| (Shen et al., 2019) | Black-box | Both |
| (Nguyen et al., 2020) | Both | Both |
| (Komkov and Petiushko, 2019) | White-box | Non-targeted |
| (Pautov et al., 2019) | White-box | Both |
| (Garofalo et al., 2018) | White-box | Non-targeted |
| (Chatzikyriakidis et al., 2019) | White-box | Targeted |
| (Kwon et al., 2019) | White-box | Targeted |
| (Dabouei et al., 2019) | White-box | Non-targeted |
| (Song et al., 2018) | White-box | Targeted |
| (Deb et al., 2019) | Black-box | Both |

*4.3.2. The Specificity*

Considering the specificity of adversarial example generation techniques, Table II represents that most attack methods are both targeted and non-targeted. Hence, the generalization is practically considered regarding this attribute. In the scenario of non-targeted attacks, which are easier to implement, Komkov and Petiushko (2019) focused on the evasion purpose of paper sticker projection on the hats, Garofalo et al. (2018) concentrated on the poisoning attack design, and Dabouei et al. (2019) prioritized the speed of their landmark-based adversarial example generation algorithm.

*4.3.3. The Transferability*

The transferability of attack methods was explored by some studies (Deb et al., 2019; Komkov and Petiushko, 2019; Zhong and Deng, 2020). Zhong and Deng (2020) explored the vulnerability of CNN-based FR models to transferable adversarial examples, spotting that feature-level attack methods are more effective and transferable than label-level ones. They observed that their proposed *DFANet* technique could enhance the transferability of existing attack methods. Komkov and Petiushko (2019) demonstrated that a paper sticker's projection on the hat with their proposed reproducible *AdvHat* method can easily confuse Face ID model LResNet100E-IR and is transferable to other Face ID models. Deb et al. (2019) verified that adversarial faces generated with their *AdvFaces* adversarial face synthesis method are model-agnostic and transferable and can evade several black-box new face matching techniques.

*4.3.4. The Perturbation*

Though universal perturbations make it easier to create adversaries in real-world applications, all reviewed attack methods in this paper demonstrated to generate image-specific perturbations. Universal perturbation generation against FR models seems to be a potential research path and is worth investing some time to avoid noise reformation any time input samples are altered (Section 6).

## 5. Defense against Adversarial Examples

As novel approaches for crafting adversarial examples are proposed, research is also directed to confront adversaries aiming to moderate their consequence on a target deep network's performance. Accordingly, several defense strategies have been defined to increase the security of at-risk FR models.

*5.1. Defense Objectives*

The objectives of defense strategies could be generally categorized into the following:

1) **Model architecture preservation** is a primary consideration taken into account when constructing any defense techniques against adversarial examples. With this objective, minimal alteration should be exerted on model architectures.
2) **Accuracy maintenance** is a primary factor considered to keep the classification outputs almost unaffected.
3) **Model speed conservation** is another factor that should not be affected during testing with the deployment of defense techniques on large datasets.

*5.2. Defense Strategies*

Generally, the defense strategies against the adversarial attacks can be divided into three categories: (1) altering the training during learning, e.g., by injecting adversarial examples into training data or incorporating altered input throughout testing, (2) changing networks, e.g., by changing the number of layers, subnetworks, loss, and activation functions, and (3) supplementing the primary model by external networks to associate in classifying unseen samples. The methodologies in the first category are not concerned with the learning models. However, the other two categories directly deal with the NNs themselves. The difference between 'changing' a network and 'supplementing' a network by external networks is that the former changes the original deep network architecture/parameters during training. Simultaneously, the latter keeps the original model intact and attaches external model(s) to it in testing. The taxonomy of the described categories is also displayed in Fig. 8. The remainder of this section is organized consistent with this taxonomy.

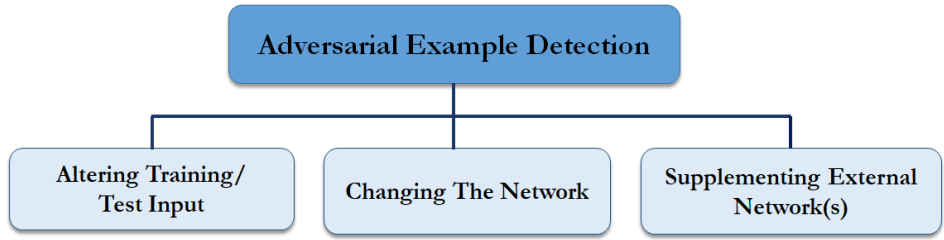

Fig. 8. A general categorization of adversarial detection methods aimed at defending FR systems against adversarial attacks.

*5.2.1. Altering Training/Test Input*

Agarwal et al. (2018) presented an efficient adversarial detection method to identify an image-agnostic universal perturbation. This method operates on (1) the pixel values and (2) the projections obtained from principal component analysis (PCA) features, as test inputs which are coupled with SVM classifier to detect perturbations. The proposed solution is considered in the first category due to flattening, hence alters the training database's images to form a row vector used either as the pixel values or dimensionally reduced vectors. The authors evaluated the effectiveness of this approach by two perturbation algorithms, universal perturbation, and a variant of it, called fast feature fool (Mopuri et al., 2017). Doing experiments with three different databases, MEDS (Founds et al., 2011), PaSC (Beveridge et al., 2013), and Multi-PIE (Gross et al., 2010), and four different DNN architectures, VGG-16, GoogLeNet, ResNet-152 (He et al., 2016), and CaffeNet (Jia et al., 2014), they showed that more straightforward approaches, such as the one proposed, can yield higher detection rates for image-agnostic adversarial perturbation. Another research (Kurnianggoro and Jo, 2019) proposed a defense strategy based on an ensemble of classification from domain transformed input data. According to this approach, input images are transformed into a grayscale format, cropped and rotated to pass the classifier, the predictions of which assembled to create the ensemble decision. The goal of this research was to discover a method that does not necessitate any retraining. On the VGGface2 dataset, experiments showed that domain transformation is useful to suppress the impact of adversarial attacks on face verification tasks.

*5.2.2. Changing the Network*

Goswami et al. (2019) proposed two defense algorithms: (1) an adversarial perturbation detection algorithm, which utilizes the CNN intermediate filter responses, and (2) a mitigation algorithm, which incorporates a specific dropout technique. In the former, authors compared the patterns of the in-between representations for original images with corresponding distorted images at each layer. They applied the differences of the two patterns to train a classifier that can categorize an unseen input as an original/distorted image. In the latter, they selectively dropped out the most affected filter responses of a CNN model, i.e., filter responses for in-between layers that reflect the most sensitivity towards noisy data to lessen the impact of adversarial noise. Subsequently, they made a comparison with unaffected filter maps. Using the VGG-Face and Light CNN networks, authors assessed the detection and mitigation algorithms according to a cross-database protocol; they performed training only with the Multi-PIE database and accomplished testing MEDS, PaSC, and MBGC (Phillips et al., 2009) databases. Across all distortions on the three databases, it was shown that the proposed detection algorithm maintains high true-positive rates even at low false-positive rates, which are desirable for the system. Also, it was observed that by discarding a certain fraction of the most affected in-between representations with the proposed mitigation algorithm, better recognition outputs could be achieved.

In another study, a blockchain security mechanism is presented to protect against FR models' attacks (Goel et al., 2019) presented. Traditional blocks of any deep learning models, such as CNNs, are converted into blocks similar to the blockchain blocks to offer fault-tolerant access in a distributed setting. In this way, tampering in one specific component alerts the entire system and easily detects 'any' probable alteration. Experiments revealed the proposed network's resilience to both the deep learning model and the biometric template, using Multi-PIE and MEDS databases.

Y. Su et al. (2019) proposed a deep *residual generative network* (*ResGN*) to clean adversarial perturbations for face verification. They suggested an innovative training framework composed of *ResGN*, VGG-Face, and FaceNet; they presented a joint of three losses: a pixel loss, a texture loss, and a verification loss, to optimize *ResGN* parameters. The VGG-Face and FaceNet networks contribute to the learning procedure by providing texture and verification losses, respectively, hence, improve the verification performance of cleaned images fundamentally. The empirical results validated the effectiveness of the proposed method on the LFW benchmark dataset. Zhong and Deng (2019) offered to recover the local smoothness of the representation space by integrating a *margin-based triplet embedding regularization* (*MTER*) term into the classification objective so that the acquired model learns to resist adversarial examples. The regularization term consists of a two-phase optimization that detects probable perturbations and punishes those using a large margin in an iterative approach. Experimental outcomes on CASIA-WebFace, VGGFace2, and MS-Celeb-1M demonstrated that the proposed method elevates network robustness against both feature-level and label-level adversarial attacks in deep FR models.

According to the concept of feature distance spaces explored in (Carrara et al., 2018), Massoli et al. (2019) proposed a detection approach based on the trajectory of internal representations, i.e., hidden layers' neuron activation, also known as deep features. They argued that the representations of adversarial inputs follow a different evolution for genuine inputs. Specifically, they collected deep features during the forward step of the target model, applied average pooling over deep features to achieve a single features vector at each selected layer, and computed the distance between each vector and the class centroid of each class at each layer, to acquire an embedding that represents the trajectory of the input image in the features space. Such a trajectory was finally fed to a binary classifier or adversarial detector. As the adversarial detector, two different architectures of a *multi-layer perceptron* (*MLP*) and a *long-short term memory* (*LSTM*) network were considered in this work. The authors conducted the experiments on the VGGFace2 dataset and the state-of-the-art Se-ResNet-50 (Cao et al., 2018). To assess the efficiency of the proposed approach, they showed the *Receiving Operating Characteristics* (*ROC*) curves from the adversarial detection considering targeted and non-targeted attacks for each architecture. They reported the *Area Under the Curve* (*AUC*) values relative to each attack. Accordingly, the AUC values were very close for the targeted attacks, while in the case of non-targeted attacks, the LSTM performance was shown to be considerably better than the MLP.

Recently, Kim et al. (2020) proposed a low-power, highly secure always-on FR processor for verification applications on mobile devices. This processor operates based on three key features of (1) a *branch net-based early stopping FR* (*BESF*) method to prevent adversarial attacks and consume low power, (2) a unified *processing element* (*PE*) for point- and depth-wise convolutions with layer fusion to reduce external memory access and (3) a *noise injection layer* (*NIL*) incorporated between bottleneck layers to make the network more robust against adversarial attacks with lower external memory access. They demonstrated that under the FGSM and PGD, BESF could result in high recognition accuracies while reducing the average power consumption significantly. They also showed that the PE reduces the external memory access, and the NIL could further lessen the FGSM and PGD attack success rates. Overall, this processor resulted in 95.5% FR accuracy in the Labeled Faces in the LFW dataset.

*5.2.3. Supplementing External Network*

Xu et al. (2017) proposed a feature squeezing strategy that moderates the search space available to an adversary by coalescing samples correspond to different feature vectors in the original space into a single sample. Adding two external models to the classifier network, they explored two feature squeezing approaches by (1) decreasing the color bit depth of each pixel and (2) spatial smoothing. Goswami et al. (2019) expressed that this approach is simple and operative for high-resolution images with detailed data; however, it may not be operational for low resolution cropped faces frequently used in FR settings. In (Goel et al., 2018), an open-source Python-based toolbox, termed as SmartBox, is proposed to benchmark the function of adversarial attack detection and mitigation algorithms against FR models. The detection approaches included in this toolbox are: 'Detection via Convolution Filter Statistics,' 'PCA-based detection,' 'Artifacts Learning' and 'Adaptive' Noise Reduction,' which are respectively considered in 'Changing the Network,' 'Altering Training/Test Input,' and 'Supplementing External Networks' defense categories. We put this study under the 'Supplementing External Networks' category since it covers the last two and hence, the majority of SmartBox detection methods.

While most of the current defense methods either assume prior knowledge of specific attacks or may not operate well on complex models due to their underlying assumptions, a new window was opened to adversarial detection techniques by leveraging the interpretability of DNNs (Tao et al., 2018). Tao et al. (2018) proposed a detection technique called *Attacks meet Interpretability* (*AmI*) in the context of FR practice. This technique features an innovative bi-directional correspondence inference amongst face attributes and internal neurons, using attribute-level mutation and neuron strengthening/weakening. More precisely, critical neurons for individual attributes are identified, and the activation values are enhanced to amplify the reasoning part of the computation. In contrast, other neurons' activation values are weakened to suppress the uninterpretable part. Employing three different datasets, VGG-Face, LFW, and CelebA, *AmI* applied to VGG-Face, with seven different kinds of attack. Extensive experiments represented that the proposed technique could successfully detect adversarial samples with a true-positive rate of 94% on average, which is significantly higher than what was achieved with the state-of-the-art reference technique, called feature squeezing (Xu et al., 2017). Similarly, the false positive rate, i.e., misclassification rate of benign inputs as malicious, of the *AmI* technique, is lower than the reference work, demonstrating its high effectiveness in this endeavor. A general overview of different adversarial example detection approaches, along with their category, is provided in Table III.

Table 3. Adversarial example detection approaches.

| Representative study | Defense strategies | Description |
| --- | --- | --- |
| (Agarwal et al., 2018) | Altering training/test input | Image pixels + PCA + SVM |
| (Kurnianggoro and Jo, 2019) | Altering training/test input | An ensemble of classification results from domain transformed (grayscale, cropped and rotated) input data |
| (Goswami et al., 2019) | Changing the network | Filter responses of CNN; dropout of filter responses |
| (Goel et al., 2019) | Changing the network | Conversion of traditional blocks of deep learning models into blocks similar to the blocks in the blockchain |
| (Y. Su et al., 2019) | Changing the network | Design of *ResGN* model + employment of a pixel loss, a texture loss, and a verification loss for parameter optimization |
| (Zhong and Deng, 2019) | Changing the network | Integration of MTER term into the classification objective for detection and punishment of perturbations |
| (Massoli et al., 2019) | Changing the network | Exploration of the adversary's evolution by tracking the trajectory of deep features representations |
| (Kim et al., 2020) | Changing the network | Design of a low-power and highly secure always-on FR processor |
| (Xu et al., 2017) | Supplementing external network(s) | Feature squeezing strategies of (1) pixel's color bit depth decreasing and (2) spatial smoothing via the addition of two external models to the classifier |
| (Goel et al., 2018) | Supplementing external network(s) | SmartBox toolbox |
| (Tao et al., 2018) | Supplementing external network(s) | Bi-directional correspondence inference amongst face attributes and internal neurons via *AmI* technique |
| (Theagarajan and Bhanu, 2020) | Supplementing external network(s) | Defending black-box FR classifiers via iterative adversarial image purifiers |

## 6. Challenges and Discussions

Although several adversarial example generation methods and defense strategies have been proposed and developed in FR's realm, various problems and challenges need to be addressed. This section summarizes the potential challenges that threaten this field. We categorize the challenges into four groups based on the literature reviewed above.

1) **Particularization/Specification of adversarial examples:** As described in this study, several image-, face-, and feature-level adversarial example generation methods have been proposed to fool FR systems; however, these methods are challenging to construct a generalized adversarial example and can only achieve good performance in certain evaluation metrics. These evaluation metrics are mainly divided into three categories: The success rate to generate adversarial examples, the robustness of the FR models, and specific attributes of the attacks, such as the perturbation amount and degree of the transferability. To explain briefly, the success rate of an attack, known as the most direct and effective evaluation criterion, is inversely proportional to the magnitude of perturbations. The robustness of FR models is related to the classification accuracy. The better the design of the FR model, the less it is vulnerable to adversarial examples. Regarding the attacks' attributes, too small perturbations on the original examples are difficult to construct adversarial examples, while too large perturbations are easily distinguished by human eyes. Therefore, a balance between constructing adversarial examples and the human visual system should be achieved. On the other hand, within a certain perturbation range, the transfer rate of adversarial examples is proportional to the magnitude of adversarial perturbations, i.e., the greater perturbations to the original example, the higher the transfer rate of the constructed adversarial examples. Taking into account these facts, the amount of perturbation to be considered on the original images, and the design of model architecture becomes critical.

Similarly, the variations in imaging conditions investigated in different works are narrower than can be encountered in practice. i.e., they are happened to be in controlled lighting, distance, etc. These conditions could be applied to some practical cases (e.g., an FR system deployed within a building). However, other practical scenarios are more challenging, needing the attacks to be tolerant of a more extensive range of imaging conditions.

These matters inhibit the defenders from designing generalized detection techniques and encourage them to propose efficient defenses against confined attacks. To overcome such challenges, a comprehensive experimental setup should be considered, possibly via scheming a standard platform as a benchmark setup setting, so that all evaluation metrics are measured simultaneously to report the efficiency of generated adversarial examples. Also, the research space should be focused more on (1) the amount of perturbation to be considered on the original images, (2) the design of FR models' architectures to be targeted, and (3) the level of transferability of generated adversarial examples. As demonstrated in Table II, the vulnerability of existing FR models to adversarial attacks in a black-box manner has been studied less, revealing the lack of transferability exploration.

2) **Instability of FR models:** Though the introduction of deep FR systems has brought benefits, it has increased the attack surface of such systems. Implementing image distortion-based adversarial attacks, for example, a substantial loss in the performance of deep learning-based systems observed, compared with the application of non-deep learning-based commercial-off-the-shelf matchers for the same evaluation data. Accordingly, the integration of only those architectures that are robust against evasion is strongly advocated. The need to develop robust models to increase adversarial examples' generalizability has been expressed in the previous paragraph, along with other influencing factors. However, this obligation is restated separately to emphasize its importance when taking steps toward generating more black-box attacks. In these circumstances, security concerns for developing more robust FR models will be raised.

3) **Deviation from the human vision system:** Adversarial attacks on vision systems exploit the fact that systems are sensitive to small changes in images to which humans are not. It will be a good idea to develop algorithms that reason images more similar to humans. In particular, those approaches that classify images based on their attributes rather than on their pixels' intensities may become more practical. Such approaches may train classifiers to recognize the presence or absence of describable aspects of visual appearances, like gender, race, age, and hair color, and extract and compare high-level visual features, or traits, of a face image that are insensitive to pose, illumination, expression, and other imaging conditions.

Profound regard to human vision physiology may open another window to research space as well. For example, the *VLA* manifested a successful implementation of physical adversarial attacks, in the design of which an attempt was made to emulate the human visual system.

4) **Image-agnostic perturbation generation:** The existing adversarial example generation methods are remarkably demonstrated to be image-agnostic, and the lack of universal perturbation generation against FR models is strongly noticed. An FR model's capability to attack different target faces simultaneously would be the by-product of generating universal perturbations, which is an essential concern in numerous studies that have been conducted in this regard.

## 7. Conclusion

This article presented a comprehensive survey in the course of adversarial attacks against intelligent deep FR systems. Despite the outstanding performance of advanced FR models, they have been vulnerable to imperceptible adversarial input images that lead them to modify their outputs entirely. This fact has opened a new window to numerous recent contributions to devise adversarial attacks and countermeasures in FR mission. This article reviewed these contributions, mainly concentrating on the most effective and inspiring works in the literature. A taxonomy of existing attack and defense methods is proposed based on different criteria. We also discussed current challenges and potential solutions in adversarial examples targeting FR models. Hope this work can shed some light on the key concepts to encourage progress in the future.